%% file: syntax.tex
\documentclass[12pt]{article}
\usepackage{chicagob}
\begin{document}
\input{defn}
\newcommand{\Ktt}{{\sf K}}

\title{Why Bother With Syntax?}
\author{Joseph Y.\ Halpern%
\thanks{Supported in part by NSF grants 
IIS-0812045, IIS-0911036, 
and CCF-1214844, AFOSR	grants FA9550-08-1-0266 and
FA9550-12-1-0040, and 
ARO	grant W911NF-09-1-0281.}\\
   Cornell University\\
   Computer Science Department\\
   Ithaca, NY 14853\\
   halpern@cs.cornell.edu\\
   http://www.cs.cornell.edu/home/halpern
}
\date{\today}
\maketitle

Rohit and I go back a long way.  We started talking about Dynamic Logic
back when I was a graduate student, when we would meet at seminars at
MIT (my advisor Albert Meyer was at MIT, although I was at Harvard, and
Rohit was then at Boston University).  Right from the beginning I
appreciated Rohit's breadth, his quick insights, his wit, and his
welcoming and gracious style.  Rohit has been interested in the
interplay between logic, philosophy, and language ever since I've known
him.  Over the years, both of us have gotten interested in game theory.
I would like to dedicate this short note, which discusses issues at the
intersection of all these areas, to him.%
\footnote{Some material in this note
appears in \cite[Section 7.2.4]{Hal31}.}

\section{Introduction}

When economists represent and reason about knowledge, they typically do
so at a {\em semantic\/} (or set-theoretic) level.  Events
correspond to sets
and the knowledge operator maps sets to sets.  On the other hand, in
the literature on reasoning about knowledge in philosophy or logic,
there is an extra layer of what may be viewed as unnecessary overhead:
syntax.  There have been papers in the economics literature
that have argued for the importance of syntax.  For example, Feinberg
\citeyear[p. 128]{Feinberg00} says ``The syntactic formalism is the more
fundamental and---intuitively---the more descriptive way to model
economic situations that involve knowledge and belief \ldots  It is fine
to use the semantic formalism, as long as what we say semantically has a
fairly clear intuitive meaning---that it can be said {\em in words}.
This amounts to saying that it can be stated syntactically.''

In this brief note, I point out some technical advantages of using
syntax.  Roughly speaking, they are the following:
\begin{enumerate}
\item Syntax allows us to
make finer distinctions than semantics; a set may be represented
by the two different expressions, and an agent may not react to these
expressions in the same
way.  Moreover, different agents may react differently to the same
expression, that is, the expression may represent different sets according to
different agents.
\item Syntax allows us to describe in a model-independent way notions
such as ``rationality''.  This enables us to identify corresponding
events (such as ``agent 1 is rational'') in two different systems.
\item The structure of the syntax provides ways
to reason and carry out proofs.  For example, many technical
results proceed by induction on the structure of formulas.  Similarly,
formal axiomatic reasoning typically takes advantage of the syntactic
structure of formulas.\index{axioms and inference rules}
\end{enumerate}

In the rest of this note, I briefly review the semantic and syntactic
approaches, explain the advantages listed above in more detail, and
point out ways in which the economics literature has not exploited
the full power of the syntactic approach thus far.  
In another paper \cite{Hal33}, I have made essentially the opposite
argument, namely, that computer scientists and logicians have not
exploited the full power of the semantic approach, pointing out that
we can often dispense with the overhead of
syntax and the need to define a $\models$ relation by working directly
with the semantics.  I believe both arguments!  This just goes to show
that each community can learn from the approaches used by the other
community.  I return to this point in the concluding section of this
paper.  

\section{Semantics vs.~Syntax: A Review}
The standard approach in the economic literature on knowledge 
starts with what is called in \cite{FHMV} an {\em Aumann structure}
$A = (\Omega,\P_1, \ldots, \P_n)$, where $\Omega$ is a set of states of
the
world and $\P_i$, $i=1, \ldots, n$ are partitions, one corresponding to
each agent.  Intuitively, worlds in the same cell of partition $i$ are
indistinguishable to agent $i$.  Given a world $\omega \in\Omega$, let
$\P_i(\omega)$ be the cell of $\P_i$ that contains $\omega$.

Knowledge operators $\Ktt_i$, $i = 1, \ldots, n$ mapping events in
(subsets of) $\Omega$ to events are defined as follows:
$$\Ktt_i(A) = \{\omega: \P_i(\omega) \subseteq A\}.$$
We read $\Ktt_i(A)$ as ``agent $i$ knows $A$''.  Intuitively, it
includes precisely those worlds $\omega$ such that $A$ holds at
all the worlds that agent $i$ cannot distinguish from $\omega$.

The logical/philosophical approach adds an extra level of indirection to
the set-theoretic approach.  The first step is to define a {\em
language\/} for reasoning about knowledge, that is, a set of well-formed
formulas.  We then associate an event with each formula.  We proceed as
follows.

We start with a set $\Phi$ of primitive propositions $p_1,
p_2, \ldots$ representing propositions of interest.  For example, $p_1$
might represent ``agent 1 is rational'' and $p_2$ might represent
``agent 2 is following the strategy of always defecting (in a game
of repeated
Prisoner's Dilemma)''.  We then close off this set under conjunction,
negation, and the {\em modal operators\/} $K_1, \ldots, K_n$.  Thus, if
$\phi$ and $\psi$ are formulas, then so are $\phi \land \psi$, $\neg
\phi$, and $K_i \phi$, for $i = 1, \ldots, n$.  We typically write $\phi
\lor \psi$ as an abbreviation for $\neg (\neg \phi \land \neg \psi)$ and
$\phi \rimp \psi$ as an abbreviation for $\neg \phi\lor \psi$.  Thus, we
can write
formulas such as $K_1 K_2 p_1 \land \neg K_2 K_1 p_2$, which could be
read as ``agent 1 knows that agent 2 knows that agent 1 is rational and
agent 2 does not know that agent 1 knows that agent 2 is following the
strategy of always defecting''.  It should be stressed that a formula
like $K_1 K_2 p_1 \land \neg K_2 K_1 p_2$ is just a string of symbols,
not a set.

A {\em Kripke structure\/} is a tuple $M = (\Omega, \P_1, \ldots, \P_n,
\pi)$, where $(\Omega,\P_1, \ldots, \P_n)$ is an Aumann structure, and
$\pi$ is an {\em interpretation}, that associates with each primitive
proposition an event in
$\Omega$.%
\footnote{In the literature, $\pi$ is often taken to associate a {\em
truth value\/} with each primitive proposition at each world; that is,
$\pi: \Phi \times \Omega \rightarrow \{{\bf true}, {\bf false}\}$.
Using this approach, we can then associate with each primitive
proposition $p$ the event $\{\omega: \pi(p,\omega) = {\bf true}\}$.
Conversely, given a mapping from primitive propositions to events, we can
construct a mapping from $\Phi \times \Omega$ to truth values.  Thus,
the two approaches are equivalent.}
We can then extend $\pi$ inductively to associate with each formula an
event as follows:
\begin{itemize}
\item $\pi(\phi \land \psi) = \pi(\phi) \inter \pi(\psi)$
\item $\pi(\neg \phi) = \overline{\pi(\phi)}$ (where $\overline{E}$
denotes the complement of $E$)
\item $\pi(K_i \phi) = \Ktt_i(\pi(\phi))$.%
\footnote{In the literature, one often sees the notation $(M,\omega)
\sat \phi$, which is read as ``the formula $\phi$ is true at world
$\omega$ in Kripke structure $M$''.  The definition of $\sat$
recapitulates that just given for $\pi$, so that $\omega \in \pi(\phi)$
iff $(M,w) \sat \phi$.}
\end{itemize}
Notice that not every subset of $\Omega$ is necessarily definable by a
formula (even if $\Omega$ is finite).  That is, for a given subset $E
\subseteq \Omega$, there may be 
no formula $\phi$ such that $\pi(\phi) = E$.  The set of events
definable by formulas form an algebra.

If all that is done with a formula is to translate it to an event, why
bother with the overhead of formulas?  Would it not just be simpler to
dispense with formulas and interpretations, and work directly with
events?  It is true that often there is no particular advantage in
working with syntax.  However, sometimes it does come in handy.  Here I
discuss the three advantages mentioned above in some more detail:
\begin{enumerate}
\item The two events $E$ and $(E \inter F) \union (E \inter
\overline{F})$ denote the same
set.  Hence, so do $\Ktt_i(E)$ and $\Ktt_i((E \inter F) \union (E \inter
\overline{F}))$.
Using the set-theoretic approach, there is no way to distinguish these
events.  Even if we modify the definition of $\Ktt$, and move to
non-partitional definitions of knowledge \cite{Bac,FHMV,Gea89,Samet87},
we still cannot distinguish these formulas.  In propositional logic,
the formula $p$ is equivalent to $(p
\land q) \lor (p \land \neg q)$.  (Of course, these formulas
were obtained
from the events by substituting $p$ for $E$, $q$ for $F$, and replacing
$\inter$, $\union$, and $\overline{\mbox{\ }}$ by $\land$, $\lor$, and
$\neg$, respectively.)  In the standard semantics defined above, the
formulas $K_ip$ and $K_i((p \land q) \lor (p \land \neg q))$ are
also equivalent.  However, an agent may not recognize that the
formulas $p$ and $(p \land q) \lor (p \land \neg q)$ are equivalent,
and may react differently to them.  There are approaches to giving semantics to
knowledge formulas that allow us to distinguish these formulas 
\cite[Chapter 9]{FHMV}.  This issue becomes important if we are
trying to model resource-bounded notions of knowledge.  
We may, for example, want to restrict the agent's knowledge to formulas
that are particularly simple, according to some notion of simplicity;
thus, it may be the case that $K_i p$ holds, while $K_i((p \land q)
\lor (p \land \neg q))$ does not.    Syntactic approaches to dealing with
\emph{awareness} \cite{FH,HR09} can capture this intuition by allowing
agent $i$ to be aware of $p$ but not of $(p \land q) \lor (p \land
\neg q)$.  This issue
also figures prominently in a recent
approach to decision theory \cite{BEH06} that takes an agent's object of
choice to be a syntactic program, which involves tests (which are
formulas).  Again, an agent's decision can, in principle, depend on the
form of the test.  A yet more general approach, where an agent's utility
function is explicitly defined on formulas in the agent's language, is
considered in \cite{BHP13}.  

A slightly more general notion of Kripke structure allows us to deal
with (at least one form of)
\emph{ambiguity}.  The standard definition of Kripke structure has a
single interpretation $\pi$.  But, in practice, agents often interpret
the same statement differently.  What one agent calls ``red'' might not be
``red'' to another agent.  It is easy to deal with this; we simply have
a different interpretation $\pi_i$ for each agent $i$ (see \cite{HK12}
for the implications of allowing such ambiguity for game theory).

\item Language allows us to describe notions in a model-independent way.
For example, the typical approach to defining rationality is to define
the event that agent $i$ is rational in a particular Aumann
structure.  But suppose we are interested in reasoning about two related
Aumann structures, $A_1$ and $A_2$, at the same time.  Perhaps each of
them has the
same set $\Omega$ of possible worlds, but different partitions.  We then
want to
discuss ``corresponding'' events in each structure, for example, events
such as ``agent 1 is rational'' or ``agent 2 is following the strategy
of always defecting''.  This is not so easy to do if we simply use the
events.  The event ``agent 1 is rational'' corresponds to different sets
in $A_1$ and $A_2$.  In the set-based approach, there is no way of
relating them.  Using syntax, an event such as ``agent 1 is rational''
would be described by a formula (the exact formula depends, of course,
on the definition of rational, a matter of some controversy; see, for
example, \cite{Binmore09,BE07,Gilboa10}).  The same formula may well
correspond to different 
events in the two structures.  For example, if the formula involves
$K_1$, then if different partitions characterize agent 1's knowledge in
$A_1$ and $A_2$, the formula would correspond to different subsets of
$\Omega$.  Although the formula describing a notion like ``agent 1 is
rational'' would correspond to different
events in different structures, we may well be able to prove general
properties of the formula that are true of all events corresponding to
the formula.  For example, if $\phi$ is the formula that says that
agent 1 is rational, we may be able to show that $\phi \rimp K_i \phi$
is {\em valid} (true in every state of every Aumann structure); this
says that if agent 1 is rational, then agent 1 knows that he is
rational.

\item There are standard examples of when the use of formulas can be
useful in proofs.  For one thing, it allows us to prove
results by induction on the structure of formulas.  Examples of such
proofs can be found in \cite{FGHV92,FHMV}.  
Secondly, the syntactic structure of a formula we would like to prove
can suggest a proof technique.  For example, a disjunction can often
be proved by cases. 
Thirdly, 
when proving that certain axioms completely characterize knowledge, the
standard proof involves building a {\em canonical model}, whose worlds
can be identified with sets of formulas \cite{Aumann99,FHMV}.  
This approach is also implicitly taken by Heifetz and Samet
\citeyear{HeSa98}.  They construct a \emph{universal type space}, a
space that contains all possible types of agents (where a \emph{type},
roughly speaking, is a complete description of an agent's beliefs about
the world and about other agents' types).  Their construction involves
certain events that they call \emph{expressions}; these expressions are
perhaps best thought of as syntactic expressions in a language of belief.
\end{enumerate}

At some level, the economics community is already aware of the
advantages of syntax, and syntactic expressions are used with varying
levels of formality in a number of papers.  To cite one example,
in \cite{BalkenborgWinter} the notion of an {\em epistemic
expression\/} is defined, that is, a function from Aumann spaces to
events in that Aumann space.  An epistemic expression such as $R_i$,
which can be thought of as representing the proposition ``agent $i$ is
rational'' then becomes a mapping from an Aumann structure to the event
consisting of all worlds where $i$ is rational.%
\footnote{Not all papers are so careful.  For example, Aumann
\citeyear{Aumann95} calls $R_i$ an event.  Of course, his intention is
quite clear.}
More complicated
epistemic expressions are then allowed, such as $\Ktt_1 \Ktt_2(R_3)$.
Perhaps not surprisingly, some proofs then proceed by induction on the
structure of event expressions.
It should be clear that event
expressions are in fact syntactic expressions, with some overloading of
notation.  Epistemic expressions are just strings of symbols.  The
$\Ktt_i$ in a
syntactic expression is not acting as a function from events to events.
It is precisely because they are strings of symbols that we can do
induction on their length.  Indeed, in \cite{FGHV92}, epistemic
expressions are introduced in the middle of a proof (under the perhaps
more appropriate name {\em (event) descriptions}, since in fact they are
used even to describe non-epistemic events), precisely to allow a proof
by induction.

\section{Discussion}

\paragraph{Expressive power:}
In the semantic approach, we can apply the knowledge operator to an
arbitrary subset of possible worlds.  In the syntactic approach,
we apply the knowledge operator to formulas.  While formulas are
associated with events, it is not necessarily the case that every event
is definable by a formula.  Indeed, one of the major issues of concern
to logicians when considering a particular syntactic formalism is its
{\em expressive power}, that is, what events are describable by formulas
in the language.

Using a more expressive formalism has both advantages and disadvantages.
The advantages are well illustrated by considering Feinberg's 
\citeyear{Feinberg00} work on characterizing the common prior assumption
(CPA) syntactically.  Feinberg considers a language that has knowledge
operators and, in addition, belief operators of the form
$p_i^\alpha(f)$, which is interpreted as ``according to agent $i$, the
probability of $f$ is at least $\alpha$''.  Feinberg does not have
common knowledge in his language, nor does his syntax allow expressions
of the form $1/2 p_i(f) + 2/3 p_i(f') \ge 1$, which can be interpreted
as ``according to agent $i$, $1/2$ of the probability of
$f$ plus $2/3$ of the probability of $f'$ is at least 1'' or, more
generally, expressions of the form $\alpha_1 p_i(f_1) + \cdots \alpha_k
p_i(f_k) \ge \beta$.  Expressions
of the latter form are allowed, for example, in \cite{FHM,FH3}.
The fact that he does not allow common knowledge causes some 
technical problems for Feinberg (which he circumvents in an elegant
way).  Linear combinations of
probability terms allow us to make certain statements about expectations
of random variables (at least, in the case of a finite state space).
Feinberg has an elegant semantic characterization of CPA in the finite
case:  Roughly speaking, he shows that CPA holds if and only if it is
not the case that there is a random variable $X$ whose expectation agent
1 judges to be positive and agent 2 judges to be negative.
Since Feinberg cannot express expectations in his language, he has to
work hard to find a syntactic characterization of CPA.  With a richer
language, it would be straightforward.

This is not meant to be a criticism of Feinberg's results.
Rather, it points out one of the
features of the syntactic approach: the issue of the exact choice of
syntax plays an important role.
There is nothing intrinsic in the syntactic
approach that prevents us from expressing notions like common knowledge
and expectation.  Rather, just as in the semantic approach, where the modeler
must decide exactly what the state space should be, in the syntactic
approach, the modeler must decide on the choice of language.  By
choosing a weaker language, certain notions become inexpressible.

Why would we want to choose a weaker language?  There are three obvious
reasons. One is aesthetic: Just as researchers judge theories on
their elegance, we have a notion of elegance for languages.  We expect
the syntax to be natural (although admittedly ``naturalness'', like
``elegance'', is in the eye of the beholder) and to avoid awkward
expressions.  A second is more practical: to the extent that we are
interested in doing inference, simpler languages typically admit simpler
inferences procedures.  This
intuition can be made formal.  There are 
numerous results characterizing the difficulty of determining
whether a formula in a given language is {\em valid} (that is, true in
every state in every structure) \cite[Chapter 3]{FHMV}.  These results
demonstrate that more complicated languages do in fact often lead to
more complex decision procedures.  However, there are also results
showing that, in this sense, there is no cost to adding certain
constructs to the language.  For example, results of \cite{FHM,FH3}
show that allowing linear combinations of probability terms does not
increase the complexity over just allowing operators such as
$p_i^\alpha$.  The third reason is that a weaker language might give us
a more appropriate level at which to study what we are most interested
in. For example, if we are interested only in qualitative beliefs ($A$
is more likely than $B$), we may not want to ``clutter up'' the language
with quantitative beliefs; that is, we may not want to be able to talk
about the exact probability of $A$ and $B$.  A simpler language lets us
focus on the key issues being examined.  Indeed, roughly speaking, we
can think of different languages as providing us with different notions
of ``sameness''; in other words, ``isomorphism'' is language-dependent, so
any result that holds ``up to isomorphism'' is really a language-dependent
result.%
\footnote{Thanks to Adam Bjorndahl for stressing this point.}

One other issue: the focus in the semantic approach has been on finding
operators that act on events.  Not all syntactic expressions correspond
in a natural way to operators.   For example, recall that in \cite{FHM},
expressions such as $\alpha_1 p_i(f_1) + \cdots \alpha_k
p_i(f_k) \ge \beta$ are considered.  This could be viewed as
corresponding to a $k$-ary operator $B^{\alpha_1, \ldots,
\alpha_k,\beta}$ such that $\omega \in B^{\alpha_1,\ldots,
\alpha_k,\beta}(E_1, \ldots, E_k)$ if $\alpha_1 p_i(E_1) + \cdots +
p_k(E_k) \ge \beta$, however, this is not such a natural operator.
It seems more natural to associate an event with this syntactic
expression directly in terms of a probability distribution on
worlds, without bothering to introduce such operators.

\paragraph{Models}  One of the advantages of the syntactic approach is
that it allows us to talk about properties without specifying a model.
This is particularly important both if we do not have a fixed model in
mind (for example, when discussing rationality, we may be interested in
general properties of rationality, independent of a particular model) or
when we do have a model in mind, but we do not wish to (or cannot)
specify it completely.  In general, a formula may be true in many
models, not just one.

To the extent that models for formulas have been considered in the
economics literature, the focus has been on one special model, the
so-called {\em canonical model} \cite{Aumnotes}.  This canonical model
has the property that if a (possibly infinite) collection of formulas
is true at some world in some model, then it is true at some
world of the canonical model.  Thus, in a sense, the canonical model can
be thought of as including all possible models.

While the canonical model is useful for various constructions, it
also has certain disadvantages.  For one thing, it is uncountable.
A related problem is that it is far too complicated to be useful as a
tool for specifying simple situations.  We can think of a formula $\phi$
as specifying a collection of structure-world pairs, namely, all the
pairs $(M,w)$ such that $(M,w) \sat \phi$.  It is well known that, at
least in the case of epistemic logic, if a formula $\phi$ is satisfiable
at all (that is, if $(M,w) \sat \phi$ for some $(M,w)$), then there is a
finite structure $M$ and a world $w$ in $M$ such that $(M,w) \sat \phi$.
Instead of focusing on an uncountable canonical structure, it is often
much easier to focus on a finite structure.

Thinking in terms of the set of models that satisfy a formula (rather
than just one canonical model) leads us to consider a number of
different issues.  In many cases we do not want to construct a model at
all.  Instead, we are interested in the logical consequences of some
properties of a model.  A formula $\phi$ is a {\em logical
consequence\/} of a collection $\Psi$ of formulas if, whenever $(M,w)
\sat \psi$ for every formula $\psi \in \Psi$, then $(M,w) \sat \phi$.
For example, we may want to know if some properties of rationality
are logical consequences of other properties of rationality; again, this
is something that is best expressed syntactically.

\section{Conclusion}
The point of this short note should be obvious: there are times when
syntax is useful, despite its overhead.
Moreover, like Moli{\`e}re's M.~Jourdain, who discovers he has
been speaking prose all his life \cite{Moliere}, game theorists have
often used syntax without realizing it (or, at least, without
acknowledging its use explicitly).  However, they have not always
taken full advantage of it.

That said, as I have pointed in \cite{Hal33}, there are times when
semantics is useful, and the overhead of syntax is not worth it.
Semantic proofs of validity are typically far easier to carry out than
their syntactic counterparts (as anyone who has tried to prove the validity of
even a simple formula like $K(\phi \land \psi) \equiv K \phi \land
K\psi$ from the axioms knows).  I believe that the finer expressive
power of syntax is particularly useful when dealing with notions of
awareness, and trying to capture how agents react differently to two
different representations of the same event.  This is something that
arguably cannot be done by a semantic approach without essentially
incorporating the syntax in the semantics.  But when considering
an approach where these issues do not arise, a semantic approach may be
the way to go.   The bottom line here is
that it is useful to have both syntactic and semantic approaches in
one's toolkit!

\paragraph{Acknowledgments:} Thanks to Adam Bjorndahl for and the two
reviewers of this paper for very useful comments.

\bibliographystyle{chicago}
\bibliography{z,refs,bghk,joe}
\end{document}

%% file: defn.tex

\newtheorem{THEOREM}{Theorem}[section]
\newenvironment{theorem}{\begin{THEOREM} \hspace{-.85em} {\bf :} }%
                        {\end{THEOREM}}
\newtheorem{LEMMA}[THEOREM]{Lemma}
\newenvironment{lemma}{\begin{LEMMA} \hspace{-.85em} {\bf :} }%
                      {\end{LEMMA}}
\newtheorem{COROLLARY}[THEOREM]{Corollary}
\newenvironment{corollary}{\begin{COROLLARY} \hspace{-.85em} {\bf :} }%
                          {\end{COROLLARY}}
\newtheorem{PROPOSITION}[THEOREM]{Proposition}
\newenvironment{proposition}{\begin{PROPOSITION} \hspace{-.85em} {\bf :} }%
                            {\end{PROPOSITION}}
\newtheorem{DEFINITION}[THEOREM]{Definition}
\newenvironment{definition}{\begin{DEFINITION} \hspace{-.85em} {\bf :} \rm}%
                            {\end{DEFINITION}}
\newtheorem{CLAIM}[THEOREM]{Claim}
\newenvironment{claim}{\begin{CLAIM} \hspace{-.85em} {\bf :} \rm}%
                            {\end{CLAIM}}
\newtheorem{EXAMPLE}[THEOREM]{Example}
\newenvironment{example}{\begin{EXAMPLE} \hspace{-.85em} {\bf :} \rm}%
                            {\end{EXAMPLE}}
\newtheorem{REMARK}[THEOREM]{Remark}
\newenvironment{remark}{\begin{REMARK} \hspace{-.85em} {\bf :} \rm}%
                            {\end{REMARK}}

\newcommand{\thm}{\begin{theorem}}
\newcommand{\lem}{\begin{lemma}}
\newcommand{\pro}{\begin{proposition}}
\newcommand{\dfn}{\begin{definition}}
\newcommand{\rem}{\begin{remark}}
\newcommand{\xam}{\begin{example}}
\newcommand{\cor}{\begin{corollary}}
\newcommand{\prf}{\noindent{\bf Proof:} }
\newcommand{\ethm}{\end{theorem}}
\newcommand{\elem}{\end{lemma}}
\newcommand{\epro}{\end{proposition}}
\newcommand{\edfn}{\bbox\end{definition}}
\newcommand{\erem}{\bbox\end{remark}}
\newcommand{\exam}{\bbox\end{example}}
\newcommand{\ecor}{\end{corollary}}
\newcommand{\eprf}{\bbox\vspace{0.1in}}
\newcommand{\beqn}{\begin{equation}}
\newcommand{\eeqn}{\end{equation}}
\newcommand{\wbox}{\mbox{$\sqcap$\llap{$\sqcup$}}}
\newcommand{\bbox}{\vrule height7pt width4pt depth1pt}
\newcommand{\qed}{\eprf}
\newcommand{\clm}{\begin{claim}}
\newcommand{\eclm}{\end{claim}}
\let\member=\in
\let\notmember=\notin
\newcommand{\sub}{_}
\def\su{^}
\newcommand{\rarrow}{\rightarrow}
\newcommand{\larrow}{\leftarrow}
\newcommand{\bolda}{{\bf a}}
\newcommand{\boldb}{{\bf b}}
\newcommand{\boldc}{{\bf c}}
\newcommand{\boldd}{{\bf d}}
\newcommand{\bolde}{{\bf e}}
\newcommand{\boldf}{{\bf f}}
\newcommand{\boldg}{{\bf g}}
\newcommand{\boldh}{{\bf h}}
\newcommand{\boldi}{{\bf i}}
\newcommand{\boldj}{{\bf j}}
\newcommand{\boldk}{{\bf k}}
\newcommand{\boldl}{{\bf l}}
\newcommand{\boldm}{{\bf m}}
\newcommand{\boldn}{{\bf n}}
\newcommand{\boldo}{{\bf o}}
\newcommand{\boldp}{{\bf p}}
\newcommand{\boldq}{{\bf q}}
\newcommand{\boldr}{{\bf r}}
\newcommand{\bolds}{{\bf s}}
\newcommand{\boldt}{{\bf t}}
\newcommand{\boldu}{{\bf u}}
\newcommand{\boldv}{{\bf v}}
\newcommand{\boldw}{{\bf w}}
\newcommand{\boldx}{{\bf x}}
\newcommand{\boldy}{{\bf y}}
\newcommand{\boldz}{{\bf z}}
\newcommand{\boldA}{{\bf A}}
\newcommand{\boldB}{{\bf B}}
\newcommand{\boldC}{{\bf C}}
\newcommand{\boldD}{{\bf D}}
\newcommand{\boldE}{{\bf E}}
\newcommand{\boldF}{{\bf F}}
\newcommand{\boldG}{{\bf G}}
\newcommand{\boldH}{{\bf H}}
\newcommand{\boldI}{{\bf I}}
\newcommand{\boldJ}{{\bf J}}
\newcommand{\boldK}{{\bf K}}
\newcommand{\boldL}{{\bf L}}
\newcommand{\boldM}{{\bf M}}
\newcommand{\boldN}{{\bf N}}
\newcommand{\boldO}{{\bf O}}
\newcommand{\boldP}{{\bf P}}
\newcommand{\boldQ}{{\bf Q}}
\newcommand{\boldR}{{\bf R}}
\newcommand{\boldS}{{\bf S}}
\newcommand{\boldT}{{\bf T}}
\newcommand{\boldU}{{\bf U}}
\newcommand{\boldV}{{\bf V}}
\newcommand{\boldW}{{\bf W}}
\newcommand{\boldX}{{\bf X}}
\newcommand{\boldY}{{\bf Y}}
\newcommand{\boldZ}{{\bf Z}}
\newcommand{\sat}{\models}
\newcommand{\dtur}{\modls}
\newcommand{\infers}{\vdash}
\newcommand{\stur}{\vdash}
\newcommand{\rimp}{\Rightarrow}
\newcommand{\limp}{\Leftarrow}
\newcommand{\dimp}{\Leftrightarrow}
\newcommand{\bor}{\bigvee}
\newcommand{\band}{\bigwedge}
\newcommand{\union}{\cup}
\newcommand{\inter}{\cap}
\newcommand{\xx}{{\bf x}}
\newcommand{\yy}{{\bf y}}
\newcommand{\uu}{{\bf u}}
\newcommand{\vv}{{\bf v}}
\newcommand{\FF}{{\bf F}}
\newcommand{\natnum}{{\sl N}}
\newcommand{\IR}{\mbox{$I\!\!R$}}
\newcommand{\IP}{\mbox{$I\!\!P$}}
\newcommand{\IN}{\mbox{$I\!\!N$}}
\newcommand{\IC}{\mbox{$C\!\!\!\!\raisebox{.75pt}{\mbox{\sqi I}}$}}
\newcommand{\marrow}{\hbox{$\rightarrow$ \hskip -10pt
                      $\rightarrow$ \hskip 3pt}}
\renewcommand{\phi}{\varphi}
\newcommand{\Circ}{\mbox{{\small $\bigcirc$}}}
\newcommand{\lt}{<}
\newcommand{\gt}{>}
\newcommand{\all}{\forall}
\newcommand{\infinity}{\infty}
\newcommand{\bc}[2]{\left( \begin{array}{c} #1 \\ #2 \end{array} \right)}
\newcommand{\cross}{\times}
\newcommand{\bigfootnote}[1]{{\footnote{\normalsize #1}}}
\newcommand{\medfootnote}[1]{{\footnote{\small #1}}}
\newcommand{\bd}{\bf}


\newcommand{\imp}{\Rightarrow}

\newcommand{\A}{{\cal A}}
\newcommand{\B}{{\cal B}}
\newcommand{\C}{{\cal C}}
\newcommand{\D}{{\cal D}}
\newcommand{\E}{{\cal E}}
\newcommand{\F}{{\cal F}}
\newcommand{\G}{{\cal G}}
\newcommand{\I}{{\cal I}}
\newcommand{\J}{{\cal J}}
\newcommand{\K}{{\cal K}}
\newcommand{\M}{{\cal M}}
\newcommand{\N}{{\cal N}}
\newcommand{\Ocal}{{\cal O}}
\newcommand{\Hcal}{{\cal H}}
\renewcommand{\P}{{\cal P}}
\newcommand{\Q}{{\cal Q}}
\newcommand{\R}{{\cal R}}
\newcommand{\T}{{\cal T}}
\newcommand{\U}{{\cal U}}
\newcommand{\V}{{\cal V}}
\newcommand{\W}{{\cal W}}
\newcommand{\X}{{\cal X}}
\newcommand{\Y}{{\cal Y}}
\newcommand{\Z}{{\cal Z}}

\newcommand{\Kone}{{\cal K}_1}
\newcommand{\abs}[1]{\left| #1\right|}
\newcommand{\set}[1]{\left\{ #1 \right\}}
\newcommand{\Ki}{{\cal K}_i}
\newcommand{\Kn}{{\cal K}_n}
\newcommand{\st}{\, \vert \,} 
\newcommand{\stc}{\, : \,} 
\newcommand{\la}{\langle}
\newcommand{\ra}{\rangle}
\newcommand{\<}{\langle}
\renewcommand{\>}{\rangle}
\newcommand{\lang}{\mbox{${\cal L}_n$}}
\newcommand{\langd}{\mbox{${\cal L}_n^D$}}

\newtheorem{nlem}{Lemma}
\newtheorem{Ob}{Observation}
\newtheorem{pps}{Proposition}
\newtheorem{defn}{Definition}
\newtheorem{crl}{Corollary}
\newtheorem{cl}{Claim}
\newcommand{\pf}{\par\noindent{\bf Proof}~~}
\newcommand{\eg}{e.g.,~}
\newcommand{\ie}{i.e.,~}
\newcommand{\vs}{vs.~}
\newcommand{\cf}{cf.~}
\newcommand{\etal}{et al.\ }
\newcommand{\resp}{resp.\ }
\newcommand{\respc}{resp.,\ }
\newcommand{\comment}[1]{\marginpar{\scriptsize\raggedright #1}}
\newcommand{\wrt}{with respect to~}
\newcommand{\re}{r.e.}
\newcommand{\nind}{\noindent}
\newcommand{\distributed}{distributed\ }
\newcommand{\bn}{\addcontentsline{toc}{section}{Notes}
\bigskip\markright{NOTES}
\section*{Notes}}
\newcommand{\Exer}{
\bigskip\markright{EXERCISES}
\section*{Exercises}}
\newcommand{\DG}{D_G}
\newcommand{\Sm}{{\rm S5}_m}
\newcommand{\Smc}{{\rm S5C}_m}
\newcommand{\Smi}{{\rm S5I}_m}
\newcommand{\Smic}{{\rm S5CI}_m}
\newcommand{\Martin}{Mart\'\i n\ }
\newcommand{\ol}{\setlength{\itemsep}{0pt}\begin{enumerate}}
\newcommand{\eol}{\end{enumerate}\setlength{\itemsep}{-\parsep}}
\newcommand{\ul}{\setlength{\itemsep}{0pt}\begin{itemize}}
\newcommand{\dl}{\setlength{\itemsep}{0pt}\begin{description}}
\newcommand{\edl}{\end{description}\setlength{\itemsep}{-\parsep}}
\newcommand{\eul}{\end{itemize}\setlength{\itemsep}{-\parsep}}
\newtheorem{fthm}{Theorem}
\newtheorem{flem}[fthm]{Lemma}
\newtheorem{fcor}[fthm]{Corollary}
\newcommand{\slidehead}[1]{
\eject
\Huge
\begin{center}
{\bf #1 }
\end{center}
\vspace{.5in}
\LARGE}

\newcommand{\subG}{_G}
\newcommand{\If}{{\bf if}}

\newcommand{\attime}{{\tt \ at\_time\ }}
\newcommand{\hatell}{\skew6\hat\ell\,}
\newcommand{\Then}{{\bf then}}
\newcommand{\Until}{{\bf until}}
\newcommand{\Else}{{\bf else}}
\newcommand{\Repeat}{{\bf repeat}}
\newcommand{\cA}{{\cal A}}
\newcommand{\cE}{{\cal E}}
\newcommand{\cF}{{\cal F}}
\newcommand{\cI}{{\cal I}}
\newcommand{\cN}{{\cal N}}
\newcommand{\cR}{{\cal R}}
\newcommand{\cS}{{\cal S}}
\newcommand{\BN}{B^{\scriptscriptstyle \cN}}
\newcommand{\BS}{B^{\scriptscriptstyle \cS}}
\newcommand{\cW}{{\cal W}}
\newcommand{\EG}{E_G}
\newcommand{\CG}{C_G}
\newcommand{\CN}{C_\cN}
\newcommand{\ES}{E_\cS}
\newcommand{\EN}{E_\cN}
\newcommand{\CS}{C_\cS}

\newcommand{\attack}{\mbox{{\it attack}}}
\newcommand{\attacking}{\mbox{{\it attacking}}}
\newcommand{\delivered}{\mbox{{\it delivered}}}
\newcommand{\exist}{\mbox{{\it exist}}}
\newcommand{\decide}{\mbox{{\it decide}}}
\newcommand{\clean}{{\it clean}}
\newcommand{\diff}{{\it diff}}
\newcommand{\Failed}{{\it failed}}
\newcommand\eqdef{=_{\rm def}}
\newcommand{\true}{{\it true}}
\newcommand{\false}{{\it false}}
\newcommand{\RAT}{\mathit{RAT}}
\newcommand{\play}{\mathit{play}}
\newcommand{\dbi}{\langle B_i \rangle}
\newcommand{\dbj}{\langle B_j \rangle}
\setcounter{secnumdepth}{2} 

\newcommand{\DN}{D_{\cN}}
\newcommand{\DS}{D_{\cS}}
\newcommand{\tyme}{{\it time}}
\newcommand{\fp}{f}

\newcommand{\Kax}{{\rm K}_n}
\newcommand{\Kaxc}{{\rm K}_n^C}
\newcommand{\Kaxd}{{\rm K}_n^D}
\newcommand{\Tax}{{\rm T}_n}
\newcommand{\Taxc}{{\rm T}_n^C}
\newcommand{\Taxd}{{\rm T}_n^D}
\newcommand{\fourax}{{\rm S4}_n}
\newcommand{\fouraxc}{{\rm S4}_n^C}
\newcommand{\fouraxd}{{\rm S4}_n^D}
\newcommand{\fiveax}{{\rm S5}_n}
\newcommand{\fiveaxc}{{\rm S5}_n^C}
\newcommand{\fiveaxd}{{\rm S5}_n^D}
\newcommand{\Dax}{{\rm KD45}_n}
\newcommand{\Daxc}{{\rm KD45}_n^C}
\newcommand{\Daxd}{{\rm KD45}_n^D}
\newcommand{\LP}{{\cal L}_n}
\newcommand{\LCP}{{\cal L}_n^C}
\newcommand{\LDP}{{\cal L}_n^D}
\newcommand{\LCDP}{{\cal L}_n^{CD}}
\newcommand{\MP}{{\cal M}_n}
\newcommand{\MPr}{{\cal M}_n^r}
\newcommand{\MPrt}{\M_n^{\mbox{\scriptsize{{\it rt}}}}}
\newcommand{\MPrst}{\M_n^{\mbox{\scriptsize{{\it rst}}}}}
\newcommand{\MPelt}{\M_n^{\mbox{\scriptsize{{\it elt}}}}}
\renewcommand{\lang}{\mbox{${\cal L}_{n} (\Phi)$}}
\renewcommand{\langd}{\mbox{${\cal L}_{n}^D (\Phi)$}}
\newcommand{\fiveaxdu}{{\rm S5}_n^{DU}}
\newcommand{\LPD}{{\cal L}_n^D}
\newcommand{\fiveaxu}{{\rm S5}_n^U}
\newcommand{\fiveaxcu}{{\rm S5}_n^{CU}}
\newcommand{\LPU}{{\cal L}^{U}_n}
\newcommand{\LPCU}{{\cal L}_n^{CU}}
\newcommand{\LDPU}{{\cal L}_n^{DU}}
\newcommand{\LCPU}{{\cal L}_n^{CU}}
\newcommand{\LPDU}{{\cal L}_n^{DU}}
\newcommand{\LPCDU}{{\cal L}_n^{\it CDU}}
\newcommand{\Cn}{\C_n}
\newcommand{\CSnp}{\I_n^{oa}(\Phi')}
\newcommand{\CSc}{\C_n^{oa}(\Phi)}
\newcommand{\Ccs}{\C_n^{oa}}
\newcommand{\CSAX}{OA$_{n,\Phi}$}
\newcommand{\CSAXN}{OA$_{n,{\Phi}}'$}
\newcommand{\untill}{U}
\newcommand{\until}{\, U \,}
\newcommand{\amp}{{\rm a.m.p.}}
\newcommand{\commentout}[1]{}
\newcommand{\msgc}[1]{ @ #1 }
\newcommand{\Camp}{{\C_n^{\it amp}}}
\newcommand{\bi}{\begin{itemize}}
\newcommand{\ei}{\end{itemize}}
\newcommand{\be}{\begin{enumerate}}
\newcommand{\ee}{\end{enumerate}}
\newcommand{\rarrowr}{\stackrel{r}{\rightarrow}}
\newcommand{\ack}{\mbox{\it ack}}
\newcommand{\Gz}{\G_0}
\newcommand{\denselist}{\itemsep 0pt\partopsep 0pt}
\def\seealso#1#2{({\em see also\/} #1), #2}
\newcommand{\cents}{\hbox{\rm \rlap{/}c}}

%% file: syntax.bbl
\begin{thebibliography}{}

\bibitem[\protect\citeauthoryear{Aumann}{Aumann}{1989}]{Aumnotes}
Aumann, R.~J. (1989).
\newblock Notes on interactive epistemology.
\newblock Cowles Foundation for Research in Economics working paper.

\bibitem[\protect\citeauthoryear{Aumann}{Aumann}{1995}]{Aumann95}
Aumann, R.~J. (1995).
\newblock Backwards induction and common knowledge of rationality.
\newblock {\em Games and Economic Behavior\/}~{\em 8}, 6--19.

\bibitem[\protect\citeauthoryear{Aumann}{Aumann}{1999}]{Aumann99}
Aumann, R.~J. (1999).
\newblock Interactive epistemology {I}: knowledge.
\newblock {\em International Journal of Game Theory\/}~{\em 28\/}(3), 263--300.

\bibitem[\protect\citeauthoryear{Bacharach}{Bacharach}{1985}]{Bac}
Bacharach, M. (1985).
\newblock Some extensions of a claim of {A}umann in an axiomatic model of
  knowledge.
\newblock {\em Journal of Economic Theory\/}~{\em 37}, 167--190.

\bibitem[\protect\citeauthoryear{Balkenborg and Winter}{Balkenborg and
  Winter}{1997}]{BalkenborgWinter}
Balkenborg, D. and E.~Winter (1997).
\newblock A necessary and sufficient epistemic condition for playing backward
  induction.
\newblock {\em Journal of Mathematical Economics\/}~{\em 27}, 325--345.

\bibitem[\protect\citeauthoryear{Binmore}{Binmore}{2009}]{Binmore09}
Binmore, K. (2009).
\newblock {\em Rational Decisions}.
\newblock Princeton, N. J.: Princeton University Press.

\bibitem[\protect\citeauthoryear{Bjorndahl, Halpern, and Pass}{Bjorndahl
  et~al.}{2013}]{BHP13}
Bjorndahl, A., J.~Y. Halpern, and R.~Pass (2013).
\newblock Language-based games.
\newblock In {\em Theoretical Aspects of Rationality and Knowledge:
  Proc.~Fourteenth Conference (TARK 2013)}, pp.\  39--48.

\bibitem[\protect\citeauthoryear{Blume and Easley}{Blume and
  Easley}{2008}]{BE07}
Blume, L. and D.~Easley (2008).
\newblock Rationality.
\newblock In L.~Blume and S.~Durlauf (Eds.), {\em The New Palgrave: A
  Dictionary of Economics}. New York: Palgrave Macmillan.

\bibitem[\protect\citeauthoryear{Blume, Easley, and Halpern}{Blume
  et~al.}{2006}]{BEH06}
Blume, L., D.~Easley, and J.~Y. Halpern (2006).
\newblock Redoing the foundations of decision theory.
\newblock In {\em Principles of Knowledge Representation and Reasoning:
  Proc.~Tenth International Conference (KR '06)}, pp.\  14--24.
\newblock A longer version, entitled ``Constructive decision theory'', can be
  found at http://www.cs.cornell.edu/home/halpern/papers/behfinal.pdf.

\bibitem[\protect\citeauthoryear{Fagin, Geanakoplos, Halpern, and Vardi}{Fagin
  et~al.}{1992}]{FGHV92}
Fagin, R., J.~Geanakoplos, J.~Y. Halpern, and M.~Y. Vardi (1992).
\newblock The expressive power of the hierarchical approach to modeling
  knowledge and common knowledge.
\newblock In {\em Theoretical Aspects of Reasoning about Knowledge:
  Proc.~Fourth Conference}, pp.\  229--244.

\bibitem[\protect\citeauthoryear{Fagin and Halpern}{Fagin and
  Halpern}{1988}]{FH}
Fagin, R. and J.~Y. Halpern (1988).
\newblock Belief, awareness, and limited reasoning.
\newblock {\em Artificial Intelligence\/}~{\em 34}, 39--76.

\bibitem[\protect\citeauthoryear{Fagin and Halpern}{Fagin and
  Halpern}{1994}]{FH3}
Fagin, R. and J.~Y. Halpern (1994).
\newblock Reasoning about knowledge and probability.
\newblock {\em Journal of the ACM\/}~{\em 41\/}(2), 340--367.

\bibitem[\protect\citeauthoryear{Fagin, Halpern, and Megiddo}{Fagin
  et~al.}{1990}]{FHM}
Fagin, R., J.~Y. Halpern, and N.~Megiddo (1990).
\newblock A logic for reasoning about probabilities.
\newblock {\em Information and Computation\/}~{\em 87\/}(1/2), 78--128.

\bibitem[\protect\citeauthoryear{Fagin, Halpern, Moses, and Vardi}{Fagin
  et~al.}{1995}]{FHMV}
Fagin, R., J.~Y. Halpern, Y.~Moses, and M.~Y. Vardi (1995).
\newblock {\em Reasoning About Knowledge}.
\newblock Cambridge, Mass.: MIT Press.
\newblock A slightly revised paperback version was published in 2003.

\bibitem[\protect\citeauthoryear{Feinberg}{Feinberg}{2000}]{Feinberg00}
Feinberg, Y. (2000).
\newblock Characterizing common priors in the form of posteriors.
\newblock {\em Journal of Economic Theory\/}~{\em 91}, 127--179.

\bibitem[\protect\citeauthoryear{Geanakoplos}{Geanakoplos}{1989}]{Gea89}
Geanakoplos, J. (1989).
\newblock Game theory without partitions, and applications to speculation and
  consensus.
\newblock Cowles Foundation Discussion Paper \#914, Yale University.

\bibitem[\protect\citeauthoryear{Gilboa}{Gilboa}{2010}]{Gilboa10}
Gilboa, I. (2010).
\newblock Questions in decision theory.
\newblock {\em Annual Reviews in Economics\/}~{\em 2}, 1--19.

\bibitem[\protect\citeauthoryear{Halpern}{Halpern}{1999}]{Hal33}
Halpern, J.~Y. (1999).
\newblock Set-theoretic completeness for epistemic and conditional logic.
\newblock {\em Annals of Mathematics and Artificial Intelligence\/}~{\em 26},
  1--27.

\bibitem[\protect\citeauthoryear{Halpern}{Halpern}{2003}]{Hal31}
Halpern, J.~Y. (2003).
\newblock {\em Reasoning About Uncertainty}.
\newblock Cambridge, Mass.: MIT Press.

\bibitem[\protect\citeauthoryear{Halpern and Kets}{Halpern and
  Kets}{2012}]{HK12}
Halpern, J.~Y. and W.~Kets (2012).
\newblock Ambiguous language and differences in beliefs.
\newblock In {\em Principles of Knowledge Representation and Reasoning:
  Proc.~Thirteenth International Conference (KR '12)}, pp.\  329--338.

\bibitem[\protect\citeauthoryear{Halpern and R\^ego}{Halpern and
  R\^ego}{2013}]{HR09}
Halpern, J.~Y. and L.~C. R\^ego (2013).
\newblock Reasoning about knowledge of unawareness revisited.
\newblock {\em Mathematical Social Sciences\/}~{\em 66\/}(2), 73--84.

\bibitem[\protect\citeauthoryear{Heifetz and Samet}{Heifetz and
  Samet}{1998}]{HeSa98}
Heifetz, A. and D.~Samet (1998).
\newblock Typology-free typology of beliefs.
\newblock {\em Journal of Economic Theory\/}~{\em 82}, 324--341.

\bibitem[\protect\citeauthoryear{Moli{\`e}re}{Moli{\`e}re}{2013}]{Moliere}
Moli{\`e}re (2013).
\newblock {\em Le Bourgeois Gentilhomme}.
\newblock Paris: Gallimard.

\bibitem[\protect\citeauthoryear{Samet}{Samet}{1990}]{Samet87}
Samet, D. (1990).
\newblock Ignoring ignorance and agreeing to disagree.
\newblock {\em International Journal of Game Theory\/}~{\em 52}, 190--207.

\end{thebibliography}
